\documentclass[journal,twoside,web]{mcomstd}
\usepackage{cite}
\usepackage{amsmath,amssymb,amsfonts}
\usepackage{algorithm}
\usepackage{algorithmic}
\usepackage{graphicx}
\usepackage[mathlines,switch]{lineno}
\usepackage{multicol,multirow}
\usepackage{marginnote}
\usepackage{stfloats}
\usepackage{mcomstd}
\usepackage{xurl}
\usepackage[hidelinks]{hyperref}
\usepackage{array,booktabs,tabularx}
\usepackage{array,tabularx}
\usepackage{comment}
\newcolumntype{L}[1]{%
  >{\hsize=#1\hsize\linewidth=\hsize
    \raggedright\arraybackslash}X}
\newcolumntype{C}[1]{%
  >{\hsize=#1\hsize\linewidth=\hsize
    \centering\arraybackslash}X}

\usepackage{array,tabularx,booktabs}

\newcolumntype{L}[1]{%
  >{\hsize=#1\hsize\linewidth=\hsize
    \raggedright\arraybackslash}X}

\newcolumntype{C}[1]{%
  >{\hsize=#1\hsize\linewidth=\hsize
    \centering\arraybackslash}X}

\usepackage{etoolbox}

\patchcmd{\biographynophoto}
  {\vskip 4\baselineskip plus 1fil minus 0\baselineskip}
  {\vskip 10pt}
  {}
  {}

\title{Where Should Models Run? A Collaborative Inference Framework for Embodied Intelligence \\in 6G}
\author{{Wanlu Sun, Zhenhui Yuan, \textit{Senior Member, IEEE}, Xiang Ji, \textit{Member, IEEE}, Yi Han, \textit{Member, IEEE}, \\Qing Li, \textit{Senior Member, IEEE}, and Gan Zheng, \textit{Fellow, IEEE}}
}

\pubid{\begin{tabular}{@{}p{30pc}@{}}\\[-2pt]\multicolumn{1}{c}{2471-2825/24/\$25.00 \copyright~2024~IEEE}\\[7pt]\end{tabular}}

\begin{document}

\maketitle

\funding{This work is supported by the EU Horizon Europe project DIRECT (Grant Agreement No.101299316).}


\begin{firstfootnote}
Wanlu Sun, Zhenhui Yuan and Gan Zheng are with the School of Engineering, University of Warwick, UK. (email: Wanlu.Sun.2@warwick.ac.uk, zhenhui.yuan@warwick.ac.uk and Gan.Zheng@warwick.ac.uk).

Xiang Ji is with the School of Information Engineering, Nanchang University, Nanchang, China. (email: tojixiang@ncu.edu.cn).

Han Yi is with the School of Information Engineering, Wuhan University of Technology, Wuhan, China. (email: hanyi@whut.edu.cn).

Qing Li is with Pengcheng Laboratory, Shenzhen, China (E-mail: liq@pcl.ac.cn).

\end{firstfootnote}

\bodyfont

\begin{abstract}
Embodied intelligence systems rely on fast and reliable inference to enable real-time perception, reasoning, and control. Recent advances in large-scale artificial intelligence (AI) models, including Large Language Models (LLMs) and Vision-Language-Action (VLA) models, have significantly enhanced the capabilities of embodied agents in complex environments. However, deploying these models remains challenging due to the limited computational resources and energy budgets of robotic platforms. A fundamental question is where inference should be executed. Collaborative inference, which partitions models across end devices, edge servers, and the cloud, has emerged as a promising solution. Yet determining the optimal partitioning point remains an open problem, as it depends on wireless network conditions, model structure, and privacy requirements. This article reviews the state of the art in collaborative inference for embodied intelligence and proposes an edge-centric framework in which the edge coordinates model partitioning and execution. Through both simulation and real-world experiments over WiFi and 5G networks, we demonstrate that the optimal partitioning strategy varies with network conditions and model structure. We further identify latency-minimizing partition points for ResNet-101, Qwen2.5-VL-7B, and OpenVLA-7B in collaborative inference between NVIDIA Jetson Nano or AGX Xavier devices and an AGX Thor edge server under varying wireless conditions. Finally, we discuss the future directions for collaborative inference in 6G-empowered embodied intelligence systems.

\end{abstract}


\section{Introduction}\label{sec:I}
Embodied Intelligence is emerging as a foundational technology for future cyber-physical systems. Recent advances in deep neural network (DNN) models, such as OpenVLA, RT-2 and Gemini Robotics, have significantly enhanced the cognitive capabilities of embodied agents, allowing them to perform complex tasks such as navigation, manipulation and planning. However, these models typically require substantial computational resources, often exceeding the capabilities of onboard processors available in embodied devices. Research in \cite{Teerapittayanon} demonstrates that collaborative inference is a promising approach, in which a DNN model is partitioned across devices, edge servers, and the cloud to overcome the limited computational resources of individual devices. Their results provide early evidence that layer-wise splitting reduces latency while enabling more scalable and resource-aware AI deployment.

Communication is a key bottleneck for distributed DNN inference across devices, edge servers, and the cloud. 3GPP TS 22.261 standard \cite{3GPP 22261} specifies that 5G system shall support split AI inference between endpoints, for instance, $<$10~ms for  AI model split on local factory robots and 10--100~ms for task offloading in remote-controlled robotics. Other challenges in collaborative inference include privacy, energy consumption, and financial cost. In many applications, such as medical robotics, sensitive data must remain within the local device or operational environment, making privacy a critical constraint. In addition, mobile robots operate under strict energy limitations, requiring a careful trade-off between computational performance and power usage. Furthermore, leveraging cloud resources for large foundation models can lead to significant operational expenses. For instance, commercial LLM services are typically billed per token; GPT-4o is priced at approximately \$2.5 per million input tokens and \$10 per million output tokens. Consequently, a robot agent generating 10 million output tokens per day could incur inference costs exceeding \$100 per day, highlighting the need for cost-aware inference placement and resource allocation strategies.

\begin{table*}[!t]
\centering
\caption{Representative DNN models for embodied intelligence and their communication KPIs and place of deployment in alignment with 3GPP TS 22.261 \cite{3GPP 22261}.}
\label{tab:combined_embodied_models}

\begingroup
\tablefont 

\renewcommand{\arraystretch}{1.12}
\renewcommand{\tabularxcolumn}[1]{m{#1}}
\setlength{\extrarowheight}{0.8pt}
\setlength{\tabcolsep}{2.3pt}

\newcolumntype{Y}[1]{>{%
  \hsize=#1\hsize
  \linewidth=\hsize
  \raggedright\arraybackslash}X}

\newcolumntype{Z}[1]{>{%
  \hsize=#1\hsize
  \linewidth=\hsize
  \centering\arraybackslash}X}

\def\modelcell#1{\shortstack[c]{#1}}

\begin{tabularx}{\textwidth}{@{}
  Y{1.10} Y{0.65} Z{1.05} Z{1.25} Z{0.92}
  Z{0.85} Z{0.85} Z{0.90} Z{1.06} Y{1.37}
@{}}

\toprule[0.8pt]

\textbf{Model} &
\textbf{Model type} &
\textbf{Organi\-zation} &
\textbf{Depth\textsuperscript{a}} &
\textbf{Params.} &
\textbf{\shortstack{Model\\size\\(MB)}} &
\textbf{Deploy\-ment} &
\textbf{\shortstack{Comm.\textsuperscript{b}\\latency\\(ms)}} &
\textbf{Download\newline latency\newline (ms)} &
\textbf{Application} \\

\midrule[0.45pt]

T5 &
Trans\-former &
Google &
\modelcell{E:6--24\\D:6--24} &
60M--\allowbreak 11B &
240--\allowbreak 44,000 &
Edge/\allowbreak Cloud &
-- &
1000 &
Language processing \\
\addlinespace[2pt]

Qwen2.5-\allowbreak VL-7B &
Trans\-former &
Alibaba &
\modelcell{ViT: 32\\LLM: 28} &
8.29B &
33,169 &
Device/\allowbreak Edge &
-- &
1000 &
Multimodal reasoning \\
\addlinespace[2pt]

NoMaD &
Diffu\-sion policy &
UC Berkeley &
4 (Trans.) &
19M &
76 &
Device/\allowbreak Edge &
12 (DL) &
1000 &
Navigation and exploration \\
\addlinespace[4pt]

MobileNet\allowbreak V3 &
CNN &
Google &
\modelcell{11/15\\(blocks)} &
2.54M--\allowbreak 5.48M &
10.2--\allowbreak 21.9 &
Device/\allowbreak Edge &
15 (UL) &
1000 &
Object recognition \\
\addlinespace[2pt]

ResNet-101 &
CNN &
\modelcell{Microsoft\\Research} &
\modelcell{101\\(conv./FC)} &
44.5M &
178 &
Edge/\allowbreak Cloud &
15 (UL) &
1000 &
Image recognition \\
\addlinespace[2pt]

YOLO26n &
CNN &
Ultralytics &
260 (modules) &
2.57M &
10.3 &
Device &
-- &
1000 &
Object detection \\
\addlinespace[2pt]

YOLO26l &
CNN &
Ultralytics &
392 (modules) &
26.3M &
105.2 &
Device &
-- &
1000 &
Object detection \\
\addlinespace[4pt]

RT-1 &
VLA &
Google Robotics &
\modelcell{MBConv:26\\Trans.:8} &
35M &
140 &
Device/\allowbreak Edge &
12 (DL) &
1000 &
Robotic manipulation \\
\addlinespace[2pt]

TinyVLA &
VLA &
Midea et al. &
Variant-dependent &
422M--\allowbreak 1.3B &
1688--\allowbreak 5200 &
Device &
-- &
1000 &
Robotic manipulation \\
\addlinespace[2pt]

OpenVLA-\allowbreak 7B &
VLA &
OpenVLA team &
\modelcell{ViT:24,27\\LLM:32} &
7.54B &
30,165 &
Edge/\allowbreak Cloud &
12 (DL) &
1000 &
Robotic manipulation \\
\addlinespace[2pt]

Gemini Robotics &
VLA &
Google DeepMind &
N/D &
N/D &
N/D &
Edge/\allowbreak Cloud &
12 (DL) &
1000 &
Generalist manipulation \\
\addlinespace[2pt]

GR00T N1.7 &
VLA &
NVIDIA &
\modelcell{32\\(action DiT)} &
3B &
12,000 &
Device/\allowbreak Edge/\allowbreak Cloud &
12 (DL) &
1000 &
Robotic manipulation \\

\bottomrule[0.8pt]
\end{tabularx}

\par\vspace{4pt}
\parbox{\textwidth}{%
  \raggedright
  \textit{Notes:}
  VLA: vision--language--action; N/D: not disclosed.
  Sizes estimate FP32 weights only
  (4 bytes/parameter; 1 MB = 10\textsuperscript{6} bytes),
  excluding activations, caches, and runtime overhead.

  \par\vspace{2pt}
  \textsuperscript{a}Depth refers to the labelled components
  (E/D: encoder/decoder), not a uniform whole-model count.
  MobileNetV3 uses TorchVision Small/Large bottleneck counts;
  YOLO26 uses unfused module counts.

  \par\vspace{2pt}
  \textsuperscript{b}UL/DL denote uplink/downlink.
  Communication references are illustrative mappings to selected
  3GPP split-inference scenarios;
  ``--'': no assigned reference.
}

\endgroup
\end{table*}

Table~\ref{tab:combined_embodied_models} illustrates representative embodied intelligence models and their deployment requirements. The table highlights the growing gap between model capability and the computational resources available on embodied devices. As foundation models continue to scale, deploying all intelligence locally becomes increasingly infeasible. Consequently, communication-aware distributed inference is emerging as a critical enabling technology for future embodied intelligence systems. Existing approaches for distributed AI inference can be broadly categorized into device-centric execution, edge-assisted inference, cloud-based inference, and collaborative device-edge-cloud computing.

Determining where models should execute is fundamentally a multi-objective optimization problem involving inference accuracy, communication quality, latency, energy consumption, privacy preservation, and economic cost. Existing solutions often optimize only a subset of these objectives and typically treat communication and inference as independent problems. In this article, we argue that future embodied intelligence systems require a unified communication--computation perspective. The main contributions of this article are threefold.

\begin{itemize}

    \item We propose an edge-centric collaborative inference framework for embodied intelligence across the device--edge--cloud continuum. The framework connects device monitoring, model compression and optimization, partition selection, and distributed execution, while accommodating communication resources and application-specific privacy requirements. It provides an architectural basis for coordinating these functions in future 6G systems.

    \item We investigate latency-minimizing device--edge partitioning through simulations and hardware experiments over a private 5G network. We examine how the preferred partition among the evaluated candidates varies across wireless conditions for models with distinct structures, relating the observed trends to computation costs and intermediate-feature sizes. The findings provide empirical guidance for partitioning policies that jointly account for model structure and wireless conditions when balancing computation against intermediate-feature transmission.

    \item We identify research opportunities in partitioning sequential and branched model structures, predicting end-to-end inference latency under changing operating conditions, and incorporating privacy constraints into partition selection. These directions connect the case-study findings to the broader framework and guide further research on adaptive, privacy-aware collaborative inference across different models.

\end{itemize}

This article focuses on inference deployment and execution rather than the distributed training of models. The case study evaluates the framework's partition-selection component; the broader orchestration functions are presented as architectural capabilities and research directions.

\section{Collaborative Inference in 6G}
\subsection{Standards}
The International Telecommunication Union (ITU) specifies a functional architecture for AI-enabled collaborative services across device, edge, and cloud environments in ITU-T Y.4509 \cite{ITU-T Y.4509}. The recommendation defines a reference model and interfaces for exchanging features and inference results to optimize distributed AI training and inference. It introduces three collaborative inference paradigms: Device Collaborative Inference (DCI), Edge Collaborative Inference (ECI), and Cloud Collaborative Inference (CCI). ITU-T F.748.20 \cite{ITU-T F.748.20} specifies a technical framework for DNN model partitioning that optimizes the placement of inference tasks. A key function of the framework is to extract the DNN model topology and estimate the computation latency of individual DNN layers, as well as the transmission latency between layers, by continuously monitoring uplink and downlink bandwidth. These estimates are then used to predict end-to-end inference latency and determine the optimal model partitioning strategy. 3GPP TR 37.817 \cite{3GPP TR 37.817} specifies how AI/machine learning (ML) model training and inference functions interact with the RAN—defining data exchange, control, and deployment principles. 3GPP TS 22.261 \cite{3GPP 22261} specifies an uplink latency requirement of less than 15 ms for splitting model of image recognition applications and a model downloading latency up to 1 second for AI/ML-enabled automotive networked systems.

\subsection{Inference across Device-Edge-Cloud}
A recent survey \cite{survey_inference} on edge intelligence in 6G networks highlights the role of collaborative inference across cloud, edge, and device tiers in enabling real-time and distributed intelligence under resource-constrained conditions. 
A number of offloading solutions have been proposed. \textit{Device-centric inference} approaches execute AI models entirely on embedded hardware. Techniques such as model compression, quantization, pruning, and knowledge distillation have enabled lightweight deployment on resource-constrained platforms. While these approaches do not rely on communications, they often suffer from limited model capacity and reduced inference accuracy \cite{Geng_survey}. \textit{Edge-assisted inference} offloads computation to nearby edge servers located at base stations or local data centers. Such approaches reduce end-to-end latency compared with cloud computing while enabling larger AI models. Recent studies have explored adaptive offloading and collaborative inference between devices and edge platforms \cite{Zeng}. However, most existing solutions assume static network conditions and optimize only latency or energy consumption. \textit{Cloud-based inference} provides virtually unlimited computational resources and supports state-of-the-art foundation models. Nevertheless, they introduce significant communication overhead, privacy concerns, and high operational costs. \textit{Collaborative device-edge-cloud inference } tackles the above limitations. Various optimization frameworks have been proposed \cite{Xu} to determine whether tasks should execute locally or be offloaded. Existing methods typically formulate the problem as latency minimization, energy minimization, or resource allocation optimization. 

\begin{figure*}
    \centering
    \includegraphics[width=0.9\linewidth]{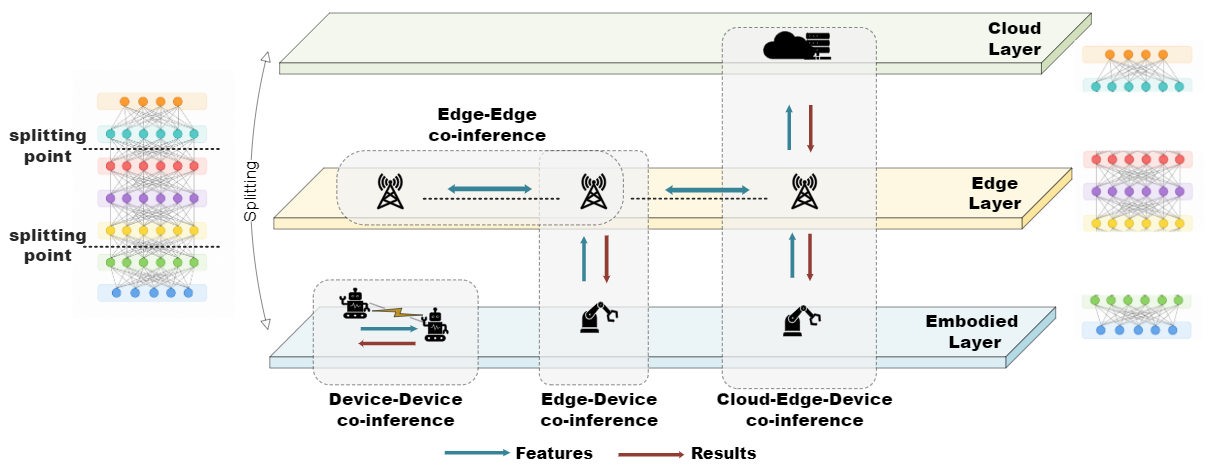}
    \caption{Illustrative architecture of collaborative inference across device, edge and cloud aligned with ITU-T Y.4509.}
    \label{fig:arch}
\end{figure*}
\subsection{Model Partitioning}
Model partitioning is a technique that splits DNN layers (or blocks of layers) across different computing platforms, such as edge devices and cloud servers. Instead of executing the entire AI model in one place, some layers run on the device close to the user, while the remaining layers are processed in the cloud. This distributed inference approach can reduce response time (latency), lower energy consumption, and enhance data privacy by limiting the amount of raw data sent over the network. Depending on available computing resources and network conditions, different partitioning strategies can be used, including single-split, multi-split, early-exit, and adaptive partitioning, which dynamically determine the most efficient point at which to divide the DNN layers. Kang et al\cite{Kang} propose a DNN partitioning framework, Neurosurgeon, that dynamically splits neural network inference between mobile devices and cloud servers to minimize latency and energy consumption while adapting to varying network and computational conditions. Results showe that their solution reduces end-to-end inference latency up to 40.7$\times$ while lowering mobile device energy consumption up to 94.7\%. They also show that the latency of individual DNN layers varies significantly. In CNN-based models, convolutional and fully connected layers account for over 90\% of the total execution time. Furthermore, the optimal partition point for CNN models is typically near the middle layers, whereas for natural language processing (NLP) models, the best partition point is usually located near either the beginning or the end of the network. Zhang et al \cite{Zhang} developes a model partitioning algorithm for Llama 2 that enables distributed deployment across heterogeneous devices, achieving up to 50\% lower latency and 2× higher throughput without compromising accuracy. Lin et al \cite{Lin_TMC}introduces a distributed learning approach that splits model training between devices and wireless edge servers to reduce the computational burden on resource-limited devices. It speeds up training and lowers communication cost by parallelising local updates and reducing the amount of data exchanged during backpropagation.

Despite recent advances in device--edge--cloud inference, existing approaches fall short of meeting the stringent requirements of embodied intelligence. Embodied agents must make real-time decisions while interacting with dynamic physical environments, requiring inference, communication, and control to operate as an integrated system. In addition, model placement decisions are usually binary, selecting either device, edge, or cloud execution.

\begin{figure*}
    \centering
    \includegraphics[width=1\linewidth]{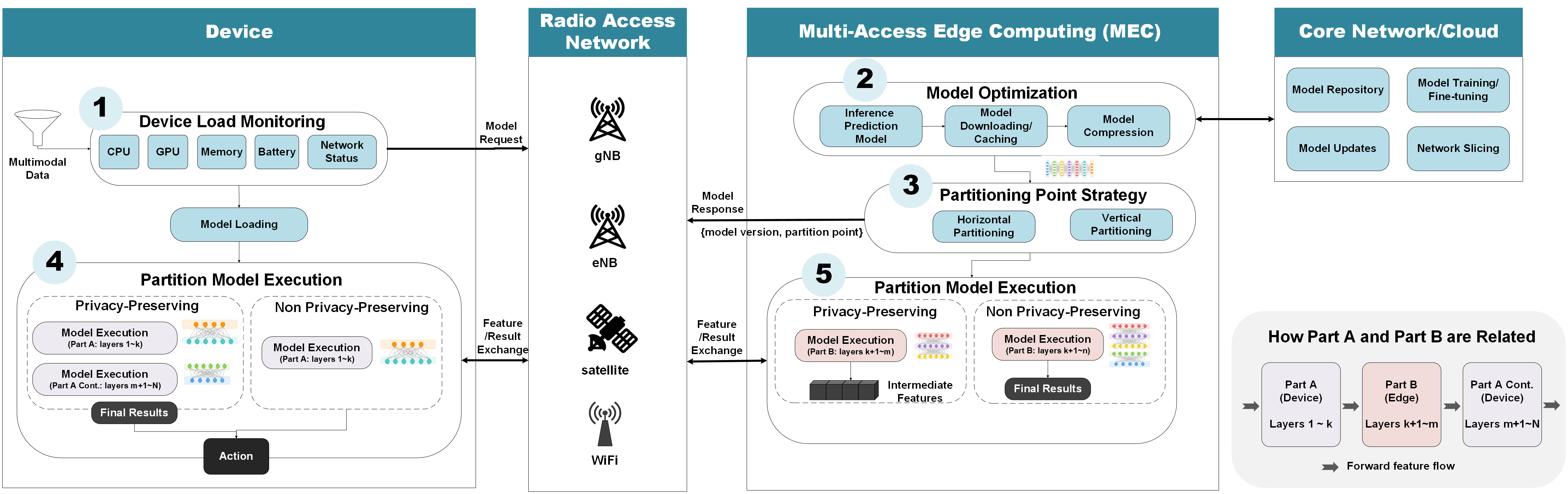}
    \caption{Functional architecture for edge-centric device–edge collaborative inference, informed by ITU-T Y.4509 and 3GPP TR 37.817. The edge server determines the model partition point(s) based on device, network, model structure and edge-server conditions. The architecture supports both two-way splitting (device–edge) and three-way splitting (device–edge–device), with the prediction head retained on the device in the latter.}
    \label{fig:block}
\end{figure*}

\section{Architecture}
Figure \ref{fig:arch} presents the proposed collaborative inference architecture. It enables execution of DNN models across the device–edge–cloud computing continuum. The optimal partitioning points are determined according to computational resources, latency requirements, model structure and network conditions, allowing inference tasks to be distributed among different computing tiers. We define four collaborative inference (co-inference) modes in alignment with ITU-T Y.4509. In \textit{Device–Device (D2D)} co-inference, model segments are executed across multiple neighboring embodied devices, improving resource utilization in infrastructure-less environments but introducing inter-device communication overhead. In \textit{Device–Edge (D2E)} co-inference, early layers are processed locally while deeper layers are offloaded to a nearby edge server, balancing communication cost and computational efficiency. \textit{Edge–Edge (E2E)} co-inference partitions the model across multiple edge servers interconnected through 5G/6G backhaul networks when a single edge node lacks sufficient resources and cloud offloading is undesirable. Finally, in \textit{Cloud–Edge–Device (C2E2D)} co-inference, large models are hierarchically distributed across cloud, edge, and end devices, enabling scalable AI services while maintaining low latency for time-critical functions.  

\subsection{Design Rationale}

The proposed framework is designed to address the latency, energy, and privacy
challenges identified in the previous section through two key design principles. i) \textit{Edge-Centric Partitioning}. Partitioning decisions are made at the edge rather than on the device primarily for scalability. The edge server has global visibility across all devices in the radio access network (RAN)—including current network congestion and its own background workload—whereas an individual device only sees its own local state. This global view lets the edge make globally-optimal (or at least system-aware) splitting decisions across many devices simultaneously, rather than each device optimizing greedily and potentially causing contention or suboptimal resource allocation at the edge. ii) \textit{Privacy-Preserving Model Splitting}. In privacy-sensitive applications, such as medical robotics, completing inference at the edge may expose sensitive outputs, including classification results, confidence scores, and probability distributions \cite{Xie_privacy}. To mitigate this risk, our framework adopts a \textit{three-way split}: the device executes the initial layers, the edge executes the intermediate, computationally
intensive layers, and the final layers of the DNN are executed back on the end device rather than at the edge. This U-shaped division, validated in \cite{Lee2023}, keeps the prediction head local, preventing sensitive outputs and task-specific or personalized model parameters from ever residing on shared edge
infrastructure. Since the final layers of modern DNNs typically account for only a small fraction of the overall inference latency and energy consumption \cite{Kang}, the three-way split preserves privacy with minimal impact on system performance, in contrast to the two-way split, which offers no such guarantee once intermediate features leave the device.

\subsection{Pipeline and Functions}

Figure~\ref{fig:block} illustrates a hierarchical co-inference pipeline that consists of five stages, spanning the device, the MEC (edge) server, and the cloud.

\textit{Stage 1 (Device: Device Load Monitoring)}: The device continuously monitors its own operating state, including CPU and GPU utilization, memory usage, remaining battery, and current network status (e.g., link quality and available bandwidth across gNB, eNB, WiFi, or satellite access). These metrics are aggregated into a status report and transmitted, together with a model request, to the MEC server via the RAN.

\textit{Stage 2 (MEC: Model Optimization)}: Upon receiving the device's request and status report, the MEC server performs three sub-tasks. First, an \textit{inference prediction} module estimates the expected end-to-end inference latency under the current device and network conditions. Second, a \textit{model download and caching} module retrieves the requested model from the cloud repository if it is not already cached locally at the edge. Third, a \textit{model compression} module applies quantization, pruning, or knowledge distillation to reduce the model footprint.

\textit{Stage 3 (MEC: Partition Strategy)}: Using the outputs of Stage 2, the MEC server runs a dynamic model partitioning algorithm to determine the optimal division of the model across the device and the edge. The stage outputs a \textit{model response} consisting of the selected model version and the corresponding partition point(s), which is transmitted back to the device via the radio access network.

\textit{Stage 4 (Device: Partition Model Execution)}: Based on the received partition point, the device first loads the assigned portion of the model and then executes the initial layers up to the partition boundary. Two
execution modes are supported: in \textit{privacy-preserving mode}, the device executes only a small number of initial layers (e.g. 1 to k layer) before offloading, consistent with a three-way split that keeps the prediction head local (see Stage 5); in \textit{non-privacy-preserving mode}, the device executes the initial layers ((e.g. 1 to k layer)) and offloads the remaining computation entirely to the MEC server, consistent with a two-way split. In both modes, this stage outputs a feature/result exchange -- the intermediate feature map -- to the MEC server.

\textit{Stage 5 (MEC: Partition Model Execution)}: The MEC server executes the remaining layers of the model, mirroring the mode selected in Stage 4. In \textit{privacy-preserving mode}, the edge executes only the intermediate layers (e.g. k+1 to m layer) and returns the resulting intermediate feature map to the device, which then completes inference locally on the final layers, so that the prediction head and its sensitive outputs never leave the device (three-way split). In \textit{non-privacy-preserving mode}, the edge executes all remaining layers (e.g. k+1~n layer) through to the output layer and returns the final prediction result directly to the device (two-way split).

Following Stage 5, the device consumes the returned feature/result and executes the corresponding \textit{Action}, triggering the associated application or robotic behavior. The cloud plays a supporting role as a global intelligence layer throughout this pipeline, responsible for model repository management, large-scale training and fine-tuning, periodic model updates, and network slicing/pricing orchestration to support diverse service requirements.

\textit{Cloud Functions}
The cloud layer provides global intelligence management and large-scale computational support. The \textit{Model Repository} maintains a collection of foundation models that can be deployed across the device--edge--cloud continuum. The \textit{Network Slicing} module further provisions differentiated communication resources tailored to the requirements of distinct embodied intelligence services, ensuring predictable performance for latency-critical, reliability-sensitive, or bandwidth-intensive workloads. Research in \cite{network_slicing} develops a 5G radio network slicing algorithm that optimizes resource allocation to enhance the performance of deep reinforcement learning applications.

\subsection{Limitations}
The proposed framework relies on stable communication, accurate inference latency prediction, and efficient model partitioning. Under highly dynamic network conditions, frequent repartitioning and model synchronization may increase communication overhead and system complexity. In addition, executing the final one or two DNN layers on the end device improves privacy but does not guarantee complete protection. While this prevents the edge from accessing application-specific outputs, the intermediate feature representations transmitted to the device may still leak information through feature inversion or representation reconstruction attacks as indicated in \cite{Xie_privacy}. Additional privacy-preserving mechanisms are therefore required for applications with stringent confidentiality requirements.

\section{Case Study: Collaborative Inference for Robotic Perception}
Building on the broader deployment framework, this section examines a central design question: how should a model be partitioned between an embodied device and an edge server to minimize end-to-end inference latency? To isolate the effects of model structure and wireless conditions within each model--platform configuration, we leave privacy-oriented three-way partitioning and joint optimization with model compression outside the scope of this study. This focused evaluation illustrates the framework's core computation--communication trade-off and provides design insights for its broader realization.

We consider a remote inspection scenario in which a robotic rover communicates with an NVIDIA Jetson AGX Thor edge server over a wireless link. The edge platform provides 128\,GB of LPDDR5X memory and up to 2070\,TFLOPS of sparse-FP4 compute. Three models are evaluated in separate experiments: ResNet-101 for visual perception, Qwen2.5-VL-7B for vision--language reasoning, and OpenVLA-7B for vision--language--action inference. Hardware validation focuses on Qwen2.5-VL-7B. To match the rover's computing resources to the local workload, we use a Jetson Nano for ResNet-101 and a Jetson AGX Xavier for the two larger models. Accordingly, comparisons focus on alternative partitions within each model--platform configuration. Alibaba Cloud serves only as an offline repository for model storage and configuration and is excluded from the real-time inference path.

\textit{Partitioning strategy and evaluation metrics.}
Each model is divided into a device-side prefix and an edge-side suffix. The rover executes the prefix and transmits its intermediate features to Thor, which completes inference. Moving the partition deeper increases
device-side computation, but reduces communication only when the resulting feature representation is smaller. 
The preferred partition therefore depends jointly on model structure, device capability, and wireless channel conditions.

We use end-to-end inference latency as the primary performance metric. In the numerical study, this latency comprises device-side computation, wireless transmission of intermediate features, and edge-side computation.
For each model, we evaluate the candidate partitioning points under different wireless channel conditions, characterized by the signal-to-noise ratio (SNR). Under each condition, the partitioning point that yields the lowest end-to-end latency is selected, and its corresponding latency components are reported. This finite-candidate evaluation examines the potential benefit of channel-dependent partitioning. It does not include online switching overhead. The hardware study additionally reports Thor-side energy per inference frame and the target-identification outcome. Rover energy is not reported because usable device-side power telemetry was unavailable in these experiments.

\subsection*{Test A: Exploration of Optimal Partitioning Points}

We conduct a simulation using an analytical simulator implemented in Python. Device and edge execution
times are estimated from partition-specific operation counts, effective computing throughputs, and fixed execution overheads. Communication latency is calculated from the transmitted feature size and a prescribed Wi-Fi SNR-to-goodput mapping, with an additional fixed network overhead. The effective computing throughputs are modeling parameters rather than the platforms' advertised peak rates.

Five representative SNR settings, 30, 25, 15, 8, and 5\,dB, are labeled Excellent, Good, Fair, Poor, and Bad, respectively. These labels describe the operating points of this study; the same SNR need not produce the same throughput or latency in Wi-Fi and 5G. For each model and channel condition, we evaluate the candidate partitions and report the one with the lowest modeled end-to-end latency in Table~\ref{tab:simulation}.

The candidate partitioning points are chosen according to each model's structure and are denoted by $k$. For ResNet-101, we evaluate $k\in\{3,4,7,8,12,18,30,31,33\}$, where each value specifies the number of residual bottleneck blocks executed on the Nano after the input stem.

For Qwen2.5-VL-7B, the partitions $k=\mathrm{S1}$, $k=\mathrm{S2}$, $k=\mathrm{S3}$, and $k=\mathrm{S4}$ correspond to splits after the first 8, 16, 24, and 32 vision Transformer blocks, respectively, with
patch embedding also executed on Xavier. At $k=\mathrm{SM}$, Xavier additionally executes the patch merger. The complete 28-layer language decoder remains on Thor.

For OpenVLA-7B, at $k=\mathrm{S1}$, $k=\mathrm{S2}$, and $k=\mathrm{S3}$, Xavier executes the patch embeddings and the first $(6,7)$, $(12,13)$, and $(18,20)$ blocks of the DINOv2 and SigLIP branches, respectively. At $k=\mathrm{S4}$, Xavier completes both visual feature-extraction branches and concatenates their outputs, while $k=\mathrm{S5}$ additionally places the multimodal projector on Xavier.
For both multimodal models, partitioning is restricted to the visual and feature-fusion components so that autoregressive decoding remains on Thor, avoiding repeated cross-device activation transfers during
token generation.

The preferred partition exhibits distinct behavior across the three models. For ResNet-101, it moves from $k=4$ under Excellent and Good
channel conditions to $k=8$ under Fair conditions and $k=31$ under Poor and Bad conditions. As SNR decreases, retaining more computation on the device becomes beneficial because the smaller intermediate representation
reduces transmission cost.

For Qwen2.5-VL-7B, $k=\mathrm{S1}$ is preferred under Excellent, Good, and Fair channel conditions, whereas $k=\mathrm{SM}$ becomes preferable under Poor and Bad conditions. The pre-merger vision blocks preserve
the feature-tensor dimensions, so moving from $k=\mathrm{S1}$ to $k=\mathrm{S4}$ adds local computation without a corresponding reduction in payload size. Executing the patch merger on the device reduces the transmitted representation, making $k=\mathrm{SM}$ advantageous at sufficiently low SNR.

For OpenVLA-7B, $k=\mathrm{S1}$ remains preferable throughout the evaluated SNR range. Within its visual branches, deeper execution does not by itself reduce the number of transmitted feature elements, and the fusion and projection stages do not guarantee a smaller payload. Under the modeled compute and communication costs, the additional device execution is therefore not offset by sufficient communication savings. These results show that the optimal split point depends on both channel conditions and model structure, explaining why a universal rule that moves the split deeper as SNR decreases is insufficient.

\begin{table*}[!t]
\centering
\caption{Simulated collaborative inference under five wireless channel conditions. For each model and condition, the table reports the latency-minimizing partition among the evaluated candidates and its
latency components.}
\label{tab:simulation}

\begingroup
\tablefont 
\setlength{\tabcolsep}{1.5pt}
\renewcommand{\arraystretch}{1.15}
\renewcommand{\tabularxcolumn}[1]{m{#1}}

\begin{tabularx}{\textwidth}{@{}
  L{1.30} L{0.80} C{0.90} C{1.15}
  C{1.00} C{1.00} C{0.85}
@{}}
\toprule

\textbf{Model} &
\textbf{SNR} &
\textbf{Partitioning point} &
\textbf{Communication latency (ms)} &
\textbf{Inference latency at device (ms)} &
\textbf{Inference latency at edge (ms)} &
\textbf{E2E latency (ms)} \\

\midrule

\multirow{5}{*}{ResNet-101}
 & Excellent & $k=4$
 & 27.69 & 40.12 & 3.52 & 71.33 \\

 & Good & $k=4$
 & 66.23 & 40.12 & 3.52 & 109.87 \\

 & Fair & $k=8$
 & 162.56 & 74.37 & 3.00 & 239.93 \\

 & Poor & $k=31$
 & 207.85 & 246.92 & 0.41 & 455.18 \\

 & Bad & $k=31$
 & 307.84 & 246.92 & 0.41 & 555.17 \\

\addlinespace[6pt]

\multirow{5}{*}{Qwen2.5-VL-7B}
 & Excellent & $k=\mathrm{S1}$
 & 22.97 & 63.87 & 433.93 & 520.77 \\

 & Good & $k=\mathrm{S1}$
 & 54.43 & 63.87 & 433.93 & 552.23 \\

 & Fair & $k=\mathrm{S1}$
 & 264.14 & 63.87 & 433.93 & 761.94 \\

 & Poor & $k=\mathrm{SM}$
 & 472.52 & 253.55 & 371.34 & 1097.41 \\

 & Bad & $k=\mathrm{SM}$
 & 701.05 & 253.55 & 371.34 & 1325.94 \\

\addlinespace[6pt]

\multirow{5}{*}{OpenVLA-7B}
 & Excellent & $k=\mathrm{S1}$
 & 37.98 & 76.06 & 1040.43 & 1154.47 \\

 & Good & $k=\mathrm{S1}$
 & 91.95 & 76.06 & 1040.43 & 1208.44 \\

 & Fair & $k=\mathrm{S1}$
 & 451.74 & 76.06 & 1040.43 & 1568.23 \\

 & Poor & $k=\mathrm{S1}$
 & 1155.18 & 76.06 & 1040.43 & 2271.67 \\

 & Bad & $k=\mathrm{S1}$
 & 1715.30 & 76.06 & 1040.43 & 2831.79 \\

\bottomrule
\end{tabularx}

\endgroup
\end{table*}

\subsection*{Test B: Collaborative Inference over a Real Wireless Network}

\begin{figure}[t]
\centering
\IfFileExists{testbed2.png}{%
  \includegraphics[width=\linewidth]{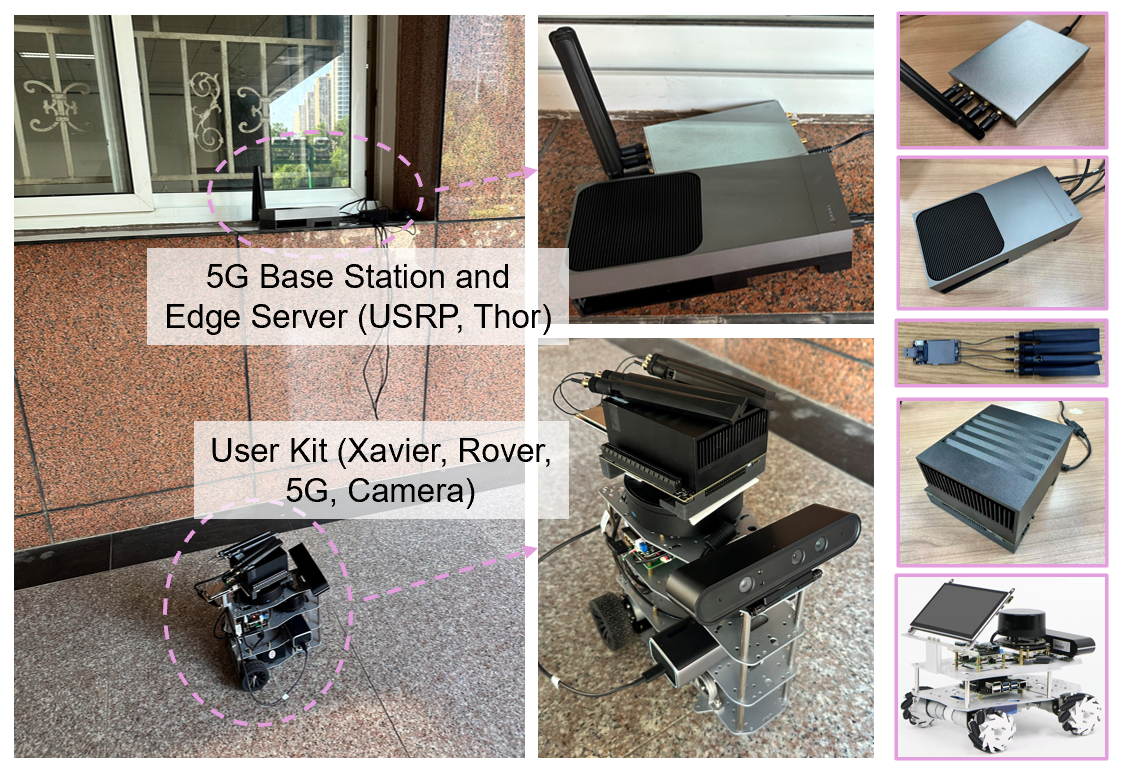}%
}{%
  \fbox{\parbox[c][24mm][c]{0.92\linewidth}{\centering
  Insert the testbed photograph here.}}%
}
\caption{Collaborative inference testbed comprising a camera-equipped robotic rover with a Jetson AGX Xavier, a private 5G link, and a Jetson AGX Thor edge server. The perception task is to identify a person wearing
a black top and carrying a white backpack.}
\label{fig:testbed}
\end{figure}

We evaluate Qwen2.5-VL-7B on the real-world testbed in Figure~\ref{fig:testbed}. A camera-equipped robotic rover carries a Jetson AGX Xavier and a 5G modem, while a Jetson AGX Thor hosts the edge inference service. 
A USRP B210-based base station provides private 5G standalone (SA) connectivity in NR band n78 with a channel bandwidth of 20\,MHz.
The rover moves away from the base station and subsequently returns, exposing the inference pipeline to varying wireless conditions. Its task is to identify a person wearing a black top and carrying a white backpack.

\begin{table*}[!t]
\centering
\caption{Hardware performance of Qwen2.5-VL-7B collaborative inference
under three measured wireless channel conditions. Bold values identify the lowest reported E2E latency within each condition.}
\label{tab:hardware}

\begingroup
\tablefont 
\setlength{\tabcolsep}{1.5pt}
\renewcommand{\arraystretch}{1.15}
\renewcommand{\tabularxcolumn}[1]{m{#1}}

\begin{tabularx}{\textwidth}{@{}
  C{0.80} C{0.95} C{1.35} C{1.10}
  C{1.10} C{1.00} C{0.95} C{0.75}
@{}}
\toprule

\textbf{Channel condition} &
\textbf{Partitioning point} &
\textbf{Communication latency (ms)} &
\textbf{Inference latency at device (ms)} &
\textbf{Inference latency at edge (ms)} &
\textbf{E2E latency (ms)} &
\textbf{Inference energy at edge (J)} &
\textbf{Target identified} \\

\midrule

\multirow{5}{*}{C1}
 & $k=\mathrm{S1}$
 & 344.96 & 78.02 & 1910.42
 & \textbf{2426.73} & 30.92 & Yes \\

 & $k=\mathrm{S2}$
 & 344.65 & 130.60 & 1872.93
 & 2447.24 & 30.74 & Yes \\

 & $k=\mathrm{S3}$
 & 357.99 & 187.06 & 1859.31
 & 2501.45 & 30.52 & Yes \\

 & $k=\mathrm{S4}$
 & 340.24 & 244.65 & 1841.87
 & 2522.61 & 30.33 & Yes \\

 & $k=\mathrm{SM}$
 & 283.66 & 247.26 & 1835.07
 & 2461.97 & 30.34 & Yes \\

\addlinespace[6pt]

\multirow{5}{*}{C2}
 & $k=\mathrm{S1}$
 & 714.65 & 79.80 & 2177.91
 & 3062.52 & 32.94 & Yes \\

 & $k=\mathrm{S2}$
 & 736.20 & 131.26 & 2115.67
 & 3082.54 & 31.36 & Yes \\

 & $k=\mathrm{S3}$
 & 724.02 & 188.26 & 2065.94
 & 3082.60 & 31.64 & Yes \\

 & $k=\mathrm{S4}$
 & 727.40 & 245.51 & 2029.70
 & 3097.76 & 31.58 & Yes \\

 & $k=\mathrm{SM}$
 & 593.58 & 247.68 & 1848.28
 & \textbf{2859.98} & 30.27 & Yes \\

\addlinespace[6pt]

\multirow{5}{*}{C3}
 & $k=\mathrm{S1}$
 & 2362.33 & 91.82 & 3415.22
 & 5957.38 & 46.45 & Yes \\

 & $k=\mathrm{S2}$
 & 2372.23 & 149.01 & 3405.15
 & 6024.48 & 46.59 & Yes \\

 & $k=\mathrm{S3}$
 & 2371.73 & 190.69 & 3371.73
 & 6034.58 & 46.25 & Yes \\

 & $k=\mathrm{S4}$
 & 2396.45 & 247.22 & 3335.19
 & 6071.61 & 45.53 & Yes \\

 & $k=\mathrm{SM}$
 & 1594.57 & 250.57 & 2872.10
 & \textbf{4834.09} & 39.55 & Yes \\

\bottomrule
\end{tabularx}

\endgroup
\end{table*}

Table~\ref{tab:hardware} compares all five candidate partitions under three measured channel conditions, denoted C1--C3 in increasing order of the communication latency observed at $k=\mathrm{S1}$. These conditions are not assigned numerical SNR intervals: communication latency depends on both channel conditions and network operation and does not uniquely determine SNR. All reported latencies and energies are expressed per inference frame.

The measured communication latency is obtained by subtracting Thor's reported server-processing time from the HTTP round-trip time and therefore includes network and protocol overhead. Device-side latency records Xavier model execution, while edge-side latency covers the remaining inference and associated server processing. Measured end-to-end latency includes these three components together with client preprocessing and serialization. These additional operations account for the difference between the measured end-to-end latency and the sum of the three tabulated latency components.

Thor-side energy is estimated from power samples collected using \texttt{tegrastats} at 100\,ms intervals. The average sampled power over the server-processing interval, expressed in watts, is multiplied by the duration of that interval, expressed in seconds, to obtain energy in joules. The resulting energy is reported per inference frame. 

Under C1, $k=\mathrm{S1}$ achieves the lowest reported end-to-end latency of 2426.73\,ms. Although $k=\mathrm{SM}$ reduces communication latency from 344.96 to 283.66\,ms, its additional Xavier computation
is not fully offset by the communication and edge-side processing savings, resulting in a total latency of 2461.97\,ms. Under C2 and C3, $k=\mathrm{SM}$ becomes preferable, reducing end-to-end latency relative to $k=\mathrm{S1}$ by 6.61\% and 18.86\%, respectively.
Under C3, switching from $k=\mathrm{S1}$ to $k=\mathrm{SM}$ reduces communication latency from 2362.33 to 1594.57\,ms and Thor-side energy from 46.45 to 39.55\,J per frame, corresponding to an energy reduction of 14.85\%. The latency-minimizing partition need not minimize edge-side energy: under C1, $k=\mathrm{S4}$ has the lowest reported edge-side energy, whereas $k=\mathrm{S1}$ has the lowest latency. All tabulated configurations report successful target identification, providing a functional check for the demonstrated scene.

The observed transition from $k=\mathrm{S1}$ to $k=\mathrm{SM}$ as communication latency increases is qualitatively consistent with Test~A. It supports the partitioning trade-off identified by the
numerical study, without establishing an identical SNR threshold or attributing every latency difference solely to channel conditions.
Simulation uses simplified computation and communication models, whereas deployment also incurs preprocessing, model-runtime overhead, and variations in wireless channel conditions. Consequently, latency
differences remain after per-frame normalization. The two sets of results should be interpreted in the context of their respective workloads and modeling assumptions: simulation characterizes performance trends under specified conditions, while hardware experiments measure the deployed inference pipeline. Their absolute difference should not be treated as prediction error under identical conditions.

\section{Future Research Opportunities}

The case study illustrates how model structure and wireless conditions jointly shape partition selection, motivating three research opportunities within the broader framework.  First, the sequential view of partitioning adopted here should be extended to more general computation graphs. For non-chain-based DNNs with parallel branches, skip connections, or multimodal fusion, partitioning must account for all transferred features and synchronization at fusion points. Future work should jointly determine branch placement and execution order. Extending partitioning into autoregressive decoders also requires considering repeated token-generation steps and the placement of cached attention states.
Second, inference latency prediction should account for preprocessing, serialization, and runtime overhead in addition to computation and transmission, as the case study demonstrates that these factors cannot be neglected. Analytical estimates can be combined with hardware profiling and online measurements to capture variations in workload and device or server load. 
Third, keeping final outputs on the device does not eliminate the risk of information leakage through intermediate features. Privacy-aware three-way partitioning should therefore define what information needs protection and what attacks the edge server could perform. Selecting the two boundaries requires balancing privacy leakage, task accuracy, local computation, and bidirectional communication latency. The cost of executing the final stage locally also needs to be assessed for each model.


\section{Conclusion}

This article presented an edge-centric framework for collaborative inference of large DNN models in 6G-oriented embodied intelligence systems. The framework integrates device and network monitoring, inference latency prediction, model retrieval and compression, partition selection, and distributed deployment and execution into a unified pipeline. It supports both device--edge and device--edge--device execution, with the latter retaining the prediction head locally to limit direct exposure of sensitive outputs.

The case study examined the framework's core partition-selection problem through simulations of three representative models and private-5G experiments with Qwen2.5-VL-7B. The results illustrate how model structure shapes the balance between computation and intermediate-feature transmission, leading to different partitioning trends as wireless conditions change. 
On the hardware testbed, selecting a different partition point reduced end-to-end latency from 5957.38 to 4834.09~ms per frame under the same channel condition, corresponding to an 18.86\% reduction. Future work should extend partitioning methods to more general computational graphs, develop reliable online latency prediction techniques, and explore privacy-aware partitioning strategies for 6G networks.


\section{Biographies}
\begin{biographynophoto}
{WANLU SUN} (Wanlu.Sun.2@warwick.ac.uk). Wanlu Sun received her B.Eng. degree from Central South University, China, in 2019 and her M.Eng. degree from Tianjin University, China, in 2022. She is currently a Ph.D. student in the Connected and Collaborative Robotics Lab at the University of Warwick, U.K., where she also serves as a teaching assistant in the School of Engineering. She is the Events and Engagement Manager for the Horizon Europe DIRECT project (2026–2030). Her research interests include resource-constrained distributed machine learning, collaborative inference for large models, energy-efficient communications, and novel methods for solving and optimizing nonlinear systems.

\end{biographynophoto}
\begin{biographynophoto}
{ZHENHUI YUAN} [IEEE Senior Member] (Zhenhui.Yuan@warwick.ac.uk). Zhenhui Yuan is an Associate Professor at the University of Warwick. He received his BEng in Software Engineering from Wuhan University, China, and PhD in Electronic Engineering from Dublin City University, Ireland. His previous roles include researcher funded by Ericsson and Enterprise Ireland, visiting scientist at UCLA, 5G Senior Research Engineer and 3GPP delegate at Huawei, and Co-Founder and CTO of RobSense Technology, which secured over RMB 15 million in venture investment. He previously served as a Senior Lecturer at Northumbria University and an Adjunct Associate Professor at Hangzhou Dianzi University. He is a telerobotics Consulting Member at University Hospitals Coventry and Warwickshire NHS Trust, an IEEE P1954 Working Group member, and Associate Editor of *IEEE Internet of Things Magazine*. He has led projects funded by EPSRC, Horizon Europe, the British Academy, and NSFC. He coordinates the Horizon Europe DIRECT project (2026–2030) and founded the VeSUS workshop series. 
\end{biographynophoto}

\begin{biographynophoto}
{XIANG JI} [IEEE Member] (tojixiang@ncu.edu.cn). Xiang Ji received his Ph.D. degree from Beijing University of Posts and Telecommunications, China. He is currently a specially appointed research fellow with the School of Information Engineering, Nanchang University, China. His research interests include transport protocols, congestion control, and network optimization. He has published first-author papers in the proceedings of IEEE INFOCOM and IJCAI and in IEEE Transactions on Cognitive Communications and Networking. He received an outstanding doctoral dissertation award from Beijing University of Posts and Telecommunications.
\end{biographynophoto}

\begin{biographynophoto}
{YI HAN} [IEEE Member] (hanyi@whut.edu.cn). Yi Han received the B.Eng. degree in software engineering from Wuhan University, China in 2010, the M.Eng. degree in telecommunication from Dublin City University, Ireland in 2011 and the Ph.D. degree from University College Dublin, Ireland in 2016. He is currently an associate professor with the School of Information Engineering,Wuhan University of Technology, China. His research interests include multimedia transmission, intelligent information perception, large model-based intelligent decision-making,and intelligent manufacturing. He has served as the principal investigator or a key technical member for multiple national-level research projects funded by the National Natural Science Foundation of China, the Major Project of the Ministry of Industry and Information Technology, and the National Key Research and Development Program of the Ministry of Science and Technology.
\end{biographynophoto}

\begin{biographynophoto}
{QING LI} [IEEE Senior Member] (liq@pcl.ac.cn). Qing Li received the B.S. degree (2008) from Dalian University of Technology, Dalian, China, and the Ph.D. degree (2013) from Tsinghua University, Beijing, China, both in computer science and technology. He is currently a full professor at Pengcheng Laboratory, Shenzhen, China. Before joining Pengcheng Laboratory, he was an associate professor at the Southern University of Science and Technology, Shenzhen, a network research scientist at Huawei (Hong Kong), and an assistant researcher at Tsinghua University. His research interests center on networking for AI, AI for networking, and intelligence delivery network, together with reliable and scalable Internet routing, software defined networking, network function virtualization, in-network caching/computing, edge computing, traffic scheduling, transmission control, and video delivery. His work aims to bridge AI systems and network infrastructure, and to architect future Internet systems that jointly optimize computation, caching, and transmission for intelligent service delivery.
\end{biographynophoto}

\begin{biographynophoto}
{GAN ZHENG} [IEEE Fellow] (Gan.Zheng@warwick.ac.\newline uk). Gan Zheng received the B.Eng. and M.Eng. degrees in electronic and information engineering from Tianjin University, Tianjin, China, in 2002 and 2004, respectively, and the Ph.D. degree in electrical and electronic engineering from The University of Hong Kong in 2008. He is currently a Professor of connected systems with the School of Engineering, University of Warwick, U.K. His research interests include machine learning for wireless communications, reconfigurable intelligent surface, fluid antennas, and edge computing. He is the first recipient of the 2013 IEEE Signal Processing Letters Best Paper Award. He also received the 2015 GLOBECOM Best Paper Award and the 2018 IEEE Technical Committee on Green Communications and Computing Best Paper Award. He was listed as a Highly Cited Researcher by Thomson Reuters/Clarivate Analytics in 2019. He serves as an Associate Editor for IEEE WIRELESS COMMUNICATIONS LETTERS, IEEE TRANSACTIONS ON COMMUNICATIONS, and IEEE TRANSACTIONS ON WIRELESS COMMUNICATIONS.
\end{biographynophoto}

\vfill

\end{document}